\documentclass{article}
\usepackage{ijcai20}

\usepackage{times}
\usepackage{soul}
\usepackage{url}
\usepackage[pagebackref,breaklinks=true]{hyperref}
\hypersetup{pdfauthor={Luca Della Libera, Vladimir Golkov, Yue Zhu, Arman Mielke, Daniel Cremers},
pdftitle={Deep Learning for 2D and 3D Rotatable Data: An Overview of Methods}}
\usepackage[utf8]{inputenc}
\usepackage[small]{caption}
\usepackage{graphicx}
\usepackage{amsmath}
\usepackage{amsthm}
\usepackage{booktabs}
\usepackage{algorithm}
\usepackage{algorithmic}
\usepackage{cmap}\usepackage[T1]{fontenc} 
\usepackage[table,x11names]{xcolor} 
\usepackage[utf8]{inputenc}
\usepackage{amssymb}
\usepackage{mathrsfs}
\usepackage{longtable}
\usepackage{tabularx}
\usepackage{breakcites}
\graphicspath{ {img/} }
\usepackage{sidecap}
\usepackage{changepage} 
\usepackage{pifont}     
\usepackage{multirow}   
\usepackage{array}      
\newcolumntype{Y}{>{\centering\arraybackslash}X} 
\newcolumntype{L}[1]{>{\RaggedRight\arraybackslash\hspace*{0pt}}p{#1}}
\usepackage{marvosym} 
\usepackage{hhline}

\usepackage{scalefnt}

\usepackage{threeparttablex}

\usepackage[overload]{textcase}

\usepackage{float} 
\usepackage{bm}
\usepackage{pifont}
\usepackage{enumitem}
\definecolor{g}{HTML}{9AFF99}
\definecolor{y}{HTML}{FCFF2F}
\definecolor{r}{HTML}{FE996B}
\sethlcolor{r}
\usepackage[tracking=true]{microtype}
\theoremstyle{definition}
\newtheorem{definition}{Definition}

\usepackage[normalem]{ulem} 

\newcommand{\bb}[1]{\textbf{#1}}
\newcommand{\mbb}[1]{\mathbb{#1}}
\newcommand{\mc}[1]{\mathcal{#1}}

\usepackage{cancel}

\newcommand{\SO}{\mathrm{SO}}
\newcommand{\SE}{\mathrm{SE}}

\usepackage{color}

\title{Deep Learning for 2D and 3D Rotatable Data: An Overview of Methods}

\author{
	Luca \MakeUppercase{Della Libera}\and
	Vladimir \MakeUppercase{Golkov}\footnote{Contact Author}\and \\
	Yue \MakeUppercase{Zhu}\and
	Arman \MakeUppercase{Mielke}\And
	Daniel \MakeUppercase{Cremers}\\
	\affiliations
	Computer Vision Group, Technical University of Munich, Germany\\
	\emails
	\{luca.della, vladimir.golkov, yue.zhu, arman.mielke, cremers\}@tum.de
}

\begin{document}

\maketitle

\thispagestyle{plain}

\begin{abstract}
	\label{abstract}
	Convolutional networks are successful due to their equivariance/invariance under translations.
	However, rotatable data such as images, volumes, shapes, or point clouds require processing with equivariance/invariance under rotations in cases where the rotational orientation of the coordinate system does not affect the meaning of the data (e.g.~object classification). On the other hand, estimation/processing of rotations is necessary in cases where rotations are important (e.g.~motion estimation).
	There has been recent progress in methods and theory in all these regards.
	Here we provide an overview of existing methods, both for 2D and 3D rotations (and translations), and identify commonalities and links between them.
\end{abstract}

\section{Introduction}
\label{sec:introduction}
Rotational and translational equivariance play an important role in image recognition tasks. Convolutional neural networks (CNNs) are translationally equivariant: the convolution of a translated image with a filter is equivalent to the convolution of the untranslated image with the same filter, followed by a translation. Unfortunately, standard CNNs do not have an analogous property for rotations.

A naive attempt to achieve rotational equivariance/invariance is data augmentation. Its major problem is that rotational equivariance in the data but not in the network architecture forces the network to learn each object orientation ``from scratch'' and hampers generalization.
Methods that achieve rotational equivariance/invariance in more advanced ways have appeared recently.

Apart from methods that are invariant under rotations of the input (i.e.~where rotation ``must not matter''), we also include examples of methods that can return rotations as output, as well as methods that use rotations as input and/or as deep features (i.e.~where rotation matters).

This work is structured as follows. In Section~\ref{sec:theory} we introduce important mathematical concepts such as equivariance and steerability. In Sections~\ref{sec:exact}--\ref{sec:learned} we present the main approaches used to achieve rotational equivariance. In Section~\ref{sec:equivariant-methods} we categorize concrete methods that use those approaches to achieve equivariance/invariance. We also categorize methods that can return a rotation as output in Section~\ref{sec:rotout} and methods that use rotations as input and/or deep features in Section~\ref{sec:input}. Finally, we draw conclusions in Section~\ref{sec:conclusions}.
The mathematical concepts (Section~\ref{sec:theory}) serve as a foundation for the best (i.e.~exact and most general) equivariant approach (Section~\ref{sec:equivariant-conv}).

\section{Formal Definitions}
\label{sec:theory}

\subsection{Equivariance}
\label{sec:equivariance}

\begin{definition}
	\label{def:equivariance}
	A function $f : \mc{X} \to \mc{Y}$ is \textbf{equivariant} under a group~$G$ (with some \emph{group actions} $\pi$ and $\psi$ of $G$ that transform $\mc{X}$ and $\mc{Y}$, respectively) if
	\begin{equation}
	\label{eq:equivariance}
	f(\pi_g[\bb{x}]) = \psi_g[f(\bb{x})] \quad \forall g\in G \quad \forall \, \bb{x}\in\mc{X},
	\end{equation}
	where $\pi_g$ is the action of $g$ on $\mc{X}$, i.e.~a transformation (for example rotation) of the input of $f$, and $\psi_g$ is the action of $g$ on $\mc{Y}$, i.e.~an ``associated'' (via the same $g$) but possibly different transformation (for example rotation of the image \emph{and} of the feature space) of the output of $f$.
	In other words, for each transformation~$\pi_g$ that modifies the input of~$f$, we know a transformation~$\psi_g$ that happens to the output of~$f$ (due to transforming the input by~$\pi_g$), without the need to know the input~$\bb{x}$.
	The usage of $g\in G$ to associate $\psi_g$ with $\pi_g$ is important for correct composition of several transformations.
	For example, if $\pi_g$ is a $180^\circ$ rotation, i.e.~$\pi_g\pi_g$ is the identity mapping, then $\psi_g\psi_g$ should also be the identity mapping.
	
	If the content of a rectangular image is rotated (and/or translated), the ``field of view'' changes, i.e.~features that used to be in the corners disappear and new features appear in the corners. This has caused some confusion as to how rectangular images can be processed in a rotation-equivariant way. The explanation is the following: An output value of the neural network is only affected if the change of features is within its receptive field and the network has not learned from data that such a feature change should be irrelevant for the output.

	Similarly, if two input features are rotated relatively to each other, an output value changes only if both input features are within its receptive field and the network has not learned that such a relative rotation should be processed equivariantly.
\end{definition}

\begin{definition}
	\label{def:same-equivariance}
	A special case of equivariance is \textbf{same\nobreakdash-equivariance} \cite{DBLP:journals/corr/DielemanFK16}, when $\psi=\pi$. In some sources, same-equivariance is called equivariance, and what we call equivariance is called covariance.
\end{definition}

\begin{definition}
	\label{def:invariance}
	A special case of equivariance is \textbf{invariance}, when 
	$\psi=\mbb{I}$, the identity.
\end{definition}

\begin{definition}
	\label{def:exact-learned}
	Equivariance is \textbf{exact} if Eq.~\eqref{eq:equivariance} holds strictly, \textbf{approximate} (for example approximated through learning) if Eq.~\eqref{eq:equivariance} holds approximatively.
\end{definition}

\subsection{Steerability}
\label{sec:steerability}

\begin{definition}
	\label{def:steerable}
	A function $f : \mc{X} \to \mc{Y}$ is \textbf{steerable} if rotated versions of $f$ can be expressed using linear combinations of a fixed set of basis functions $h_j$ for $j=1,\dots,M$, that is:
	\begin{equation}
	\label{eq:steerability}
	f(\pi[\mathbf{x}]) = \sum\limits_{j=1}^M k_j(\pi)h_j(\mathbf{x}),
	\end{equation}
	where $\pi$ is a rotation and $k_j$ are complex-valued rotation\nobreakdash-dependent \emph{steering factors}.
\end{definition}
For example, if we consider a standard non-normalized 2D Gaussian $G(x,y) = e^{-(x^2+y^2)}$, its first derivative $G_x(x,y) = \frac{\partial G}{\partial x}(x,y)$ in the $x$ direction can be steered at an arbitrary orientation~$\theta$ through a linear combination of $G_x^{0^{\circ}}(x,y) = G_x(x,y) = -2xe^{-(x^2+y^2)}$ and $G_x^{90^{\circ}}(x,y) = G_x(\pi^{90^\circ}[x,y]) = -2ye^{-(x^2+y^2)}$:
\begin{align}
\label{eq:gaussian}
G_x^{\theta}(x,y) &= G_x(\pi^\theta[x,y]) \nonumber\\ &=\cos(\theta)G_x^{0^{\circ}}(x,y) + \sin(\theta)G_x^{90^{\circ}}(x,y).
\end{align}
A visualization of the case $\theta=30^\circ$ looks as follows:
\begin{align}
\label{eq:gaussian_steered}
\underbrace{G_x^{30^\circ}(x,y)}_{\includegraphics{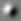}}
&= \underbrace{\cos(30^\circ)}_{\sim 0.87}\underbrace{G_x^{0^\circ}(x,y)}_{\includegraphics{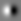}} + \underbrace{\sin(30^\circ)}_{0.5}\underbrace{G_x^{90^\circ}(x,y)}_{\includegraphics{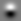}}.
\end{align}
A useful consequence of steerability is that convolution of an image with basis filters $h_j$ is rotationally equivariant.
The mapping $\psi$ in Eq.~\eqref{eq:equivariance} in this case corresponds to a linear combination of the feature maps.
As a side note, the mapping $\psi$ can also be a certain kind of linear combinations even if the filters are not a basis of a steerable filter, see Section~\ref{sec:equivariant-conv}. 

A \textbf{harmonic function} $f$ is a twice continuously differentiable function that satisfies Laplace's equation, i.e.~$\nabla^2f = 0$.
Circular harmonics and spherical harmonics are defined on the circle and the sphere, respectively, and are similar to the Fourier series (i.e.~sinusoids with different frequencies).

2D or 3D rotational equivariance can be hardwired in the network architecture by restricting the filters' angular component to belong to the circular harmonic or spherical harmonic family, respectively. 
The proof utilizes steerability properties of such filter banks.
The radial profile of these filters on the other hand can be learned. There are various techniques to parameterize the radial profile (with learnable parameters).

\subsection{Group Convolution}
\label{sec:group-conv}
\begin{definition}
	The \textbf{group convolution} \cite{DBLP:journals/corr/CohenW16,DBLP:journals/corr/abs-1711-06721} between a feature map~$\mathbf{F}$ and a filter~$\mathbf{W}$ is defined as
	\begin{equation}
	\label{eq:group-conv}
	(\mathbf{F} \star_G \mathbf{W})(\mathbf{x}) = \int_{g \in G} \mathbf{F}(\phi_g [\bm\eta]) \mathbf{W}(\phi_g^{-1} [\mathbf{\mathbf{x}}]) \,\mathrm{d}g,
	\end{equation}
	where $G$ is a group, $g \in G$ is a group element with group action $\phi_g$, and $\bm\eta$ is typically a canonical element in the domain of $\mathbf{F}$ (e.g.~the origin if the domain is $\mathbb{R}^n$).
The group convolution can be shown \cite{DBLP:journals/corr/abs-1709-01889,Kondor2018OnTG}
to be equivariant.
The ordinary convolution is a special case of the group convolution.
\end{definition}

\section{Approaches that Guarantee Exact Rotational Equivariance}
\label{sec:exact}
In this section we list and briefly discuss the approaches used to achieve exact rotational equivariance/invariance. The state\nobreakdash-of\nobreakdash-the\nobreakdash-art approach is described in Section~\ref{sec:equivariant-conv}.

\subsection{Hardwired Pose Normalization}
\label{sec:pose-norm-hardwired}
A basic approach to address the problem of rotational invariance consists in trying to ``erase'' the effect of rotations by reverting the input to a canonical pose by \emph{hardwiring} a reversion function such as PCA.
Problems are: small noise can strongly influence the result especially for objects with symmetries; learned low-level feature detectors do not generalize to other orientations.

\subsection{Handcrafted Features}
\label{sec:handcrafted}
Extractors of simple rotationally invariant features can be handcrafted rather than learned.
Examples include features based on distances between pairs of points ($\SE(n)$-invariant), and/or between each point and the origin (invariant under rotations around the origin).
Handcrafted feature extractors are not trained, i.e.~not optimal.

\subsection{General Linear Equivariant Mappings and Equivariant Nonlinearities}
\label{sec:equivariant-conv}
In many situations, the best approach are the most general methods that guarantee equivariance \cite{Kondor2018OnTG,DBLP:journals/corr/abs-1807-02547,DBLP:journals/corr/abs-1811-02017}. Several formulations are available.

When using so-called \emph{irreducible representations} of the rotation group, the equivariant mapping corresponds to the usage of steerable filters (see Section~\ref{sec:steerability}).
This enables equivariance under all infinitely many rotation angles, but pointwise nonlinearities such as ReLU are not equivariant in this basis.
Only other, special nonlinearities are equivariant. (See Section~3.3 by \cite{DBLP:journals/corr/abs-1811-02017} for an overview of equivariant nonlinearities, equivariant batch normalization, and equivariant residual learning.)

On the other hand, when using \emph{regular representations}, the mapping corresponds to group convolution, Eq.~\eqref{eq:group-conv}, i.e.~the usage of a finite number of rotated versions of arbitrary filters.
This enables equivariance only under a discrete subgroup (e.g.~$45^{\circ}$ rotations in a plane) of the rotation group, but pointwise nonlinearities like ReLU are equivariant.

For 2D rotations, results are best in practice when group convolutions with small rotation angles and pointwise nonlinearities are used \cite{weiler2019general}.

For the most common purposes, this is arguably the best of all approaches: it guarantees exact equivariance (unlike Section~\ref{sec:learned}), without an obligation to use untrainable feature extractors (unlike Section~\ref{sec:handcrafted}) or unstable pose normalization (unlike Section~\ref{sec:pose-norm-hardwired}).

Note that the exactness of the equivariance is slightly reduced when data are discretized to a pixel/voxel grid. Deeper layers might further amplify the impact of the angle between the grid and the object on the features. A part of this ``missing part of equivariance'' can be \emph{learned} from training data (see Section~\ref{sec:augmentation}). If the trained network has such ``partially learned'' equivariance, it is not obvious how it achieves it, i.e.~the group action $\psi$ of intermediate layers is not known, unlike in the case of exact equivariance.

\section{Approaches to Learn Approximate Rotational Equivariance}
\label{sec:learned}
Various approaches exist that facilitate the learning of approximated (inexact) rotational equivariance.
Note that for datasets where exact equivariance is appropriate, approaches that provide exact equivariance (Section~\ref{sec:equivariant-conv}) usually work better.

\subsection{Data Augmentation}
\label{sec:augmentation}
Data augmentation (i.e.~random rotations of samples during training) is the most naive approach to deal with rotational equivariance.
Such rotational equivariance in the training data but not in the network architecture forces the network to learn to recognize each orientation of each object part ``from scratch'' and hampers generalization.

\subsection{Learned Pose Normalization}
\label{sec:pose-norm-learned}
Instead of \emph{hardwiring} a pose normalization function as described in Section~\ref{sec:pose-norm-hardwired}, it is possible to force or encourage the network to \emph{learn} a reversion function directly from the training data. As an example of the ``encourage'' case, in spatial transformer networks \cite{Jaderberg:2015:STN:2969442.2969465}, learning a pose normalization is a facilitated but not a guaranteed side effect of learning to classify.

\subsection{Soft Constraints}
\label{sec:soft-constraints}
Another approach to let the network learn rotational equivariance/invariance is to introduce additional \emph{soft constraints}, which are typically expressed by auxiliary loss functions that are added to the main loss function. For example, a similarity loss
\cite{Coors2018VISAPP} can be defined,
which penalizes large distances between the predictions or feature embeddings of rotated copies of the input that are simultaneously fed into separate streams of a siamese network.

The advantage of this approach is the ease of implementation. Furthermore, it can be used in combination with other approaches that provide non-exact equivariance (e.g.~pose normalization) in order to enhance it. The disadvantage is that equivariance/invariance is only approximative. The quality of the approximation depends on the loss formula, training data, network architecture and optimization algorithm.

\subsection{Deformable Convolution}
\label{sec:deformable-conv}
The \emph{deformable convolution} \cite{DBLP:journals/corr/DaiQXLZHW17} augments a CNN's capability of modeling geometric transformations. Input-dependent offsets are added to the sampling locations of the standard convolution.
The offsets are computed by applying an additional convolutional layer over the same input feature map. Bilinear image interpolation is used due to non-integer offsets.
The advantage of this approach is that it can learn to handle very general transformations such as rotation, scaling, and deformation, if training data encourage this. The disadvantage is that there is no guarantee of equivariance.

\section{Overview of Methods}
\label{sec:overview}
In this section we list and categorize deep learning methods for handling rotatable data and rotations. This includes methods that are equivariant under rotations of the input, methods that output a rotation, and methods that use rotations as inputs and/or deep features.

\subsection{Equivariance under Rotations of the Input}
\label{sec:equivariant-methods}

Methods that are equivariant under rotations of the input are categorized in \autoref{tbl:equiv:2d} (for 2D rotations) and \autoref{tbl:equiv:3d} (for 3D rotations) according to the following criteria:

\begin{itemize}
	\phantomsection\label{par:input}\item{Input:}
	\begin{itemize}
		\item Pixel grid: a grid representation of 2D image data
		\item Voxel grid: a grid representation of 3D volumetric data
		\item Point cloud: a set of 3D point coordinates
		\item Spherical signal: a function defined on the sphere
		\item Polygon mesh: a collection of vertices, edges, and faces that describes a surface consisting of polygons
		\item dMRI (6D): six-dimensional diffusion-weighted magnetic resonance images~\cite{mueller2021}
	\end{itemize}

	\phantomsection\label{par:approach}\item{Approach: see Definition~\ref{def:exact-learned} and Sections~\ref{sec:exact}--\ref{sec:learned}}
	
	\phantomsection\label{par:property}\item{Property: equivariance (Definition~\ref{def:equivariance}) or invariance (Definition~\ref{def:invariance})}

	\phantomsection\label{par:group}\item{Group:}
	\begin{itemize}
		\item $\SO(2)$: the group of 2D rotations
		\item $\SE(2)$: the group of 2D rigid-body motions
		\item $\SO(3)$: the group of 3D rotations
		\item $\SE(3)$: the group of 3D rigid-body motions
	\end{itemize}

	\phantomsection\label{par:cardinality}\item{Cardinality: continuous (entire group) or discretized to specific angles}

	\label{par:type}
\end{itemize}

\subsection{Rotations as Output}
\label{sec:rotout}

Examples of deep learning methods that output a (3D) rotation are categorized in \autoref{tbl:output:gt} and \autoref{tbl:output:nogt} according to the following characteristics:

\begin{itemize}
	\item \hypertarget{sec:rotout:input}{Input to the network that outputs the rotation, and according rotation-prediction task:}
	\begin{itemize}
		\item Image.
		The task is to estimate the orientation of a depicted object relative to the camera.
		\item Cropped stereo image:
		A pair of images is taken at the same time from two cameras that are close together and point in the same direction.
		The images are cropped in a predetermined fashion.
		Each pair of cropped images constitutes one input.
		The position of the cameras relative to each other is fixed. The task is to estimate the orientation of a depicted object relative to the cameras.
	
	\begin{table}[t!]
\begin{threeparttable}[b]
\setlength\arrayrulewidth{.1pt}
    \footnotesize
	 \begin{tabularx}{\columnwidth}{@{}|>{\raggedright\arraybackslash}X|L{1.17cm}|L{0.69cm}|L{0.80cm}|L{1.58cm}|@{}}
        \hline
        Method & \hyperref[par:approach]{Approach} & \hyperref[par:property]{Property} & \hyperref[par:group]{Group} & \hyperref[par:cardinality]{Cardinality} \\
        \hline \hline
        Many & 
            \hl{Learned} \newline\tiny (Data \newline augmentation) & 
            * & * & *  \\ \hline
        Spatial Transformer Networks \cite{Jaderberg:2015:STN:2969442.2969465} &
            \hl{Learned} \newline\tiny (Learned~pose \mbox{normalization)} & Invariance & $\SE(2)$ & Continuous  \\ \hline
        Cyclic Symmetry in CNNs \cite{DBLP:journals/corr/DielemanFK16} &
            Exact &
            Equivar. & $\SE(2)$ & \hl{Discretized} \newline ($90^{\circ}$ angles)
            \\ \hline
        Group Equivariant CNNs \cite{DBLP:journals/corr/CohenW16}  &
            Exact & Equivar. & $\SE(2)$ & \hl{Discretized} \newline ($90^\circ$ angles) \\ \hline
        Harmonic Networks \newline \cite{DBLP:journals/corr/WorrallGTB16} &
            Exact &
            Equivar. & $\SE(2)$ & Continuous \\ \hline
        Vector Field Networks \newline\cite{DBLP:journals/corr/GonzalezVKT16} &
            Exact &
            Equivar. & $\SE(2)$ & \hl{Discretized} \newline (any angle)
            \\ \hline
        Oriented~Response~Networks~\mbox{\cite{DBLP:journals/corr/ZhouYQJ17}} &
            Exact &
            Equivar. & $\SE(2)$ & \hl{Discretized} \newline (any angle)
            \\ \hline
        Deformable CNNs \newline\cite{DBLP:journals/corr/DaiQXLZHW17} &
            \hl{Learned} \newline\tiny (Deformable \newline convolution) & Equivar. & $\SE(2)$ & Continuous  \\ \hline
        Polar Transformer Networks \newline\cite{DBLP:journals/corr/abs-1709-01889} 
            & \hl{Learned} \newline\tiny (Learned~pose \mbox{normalization)} & Equivariance & $\SE(2)$ & Continuous  \\ \hline
        Steerable Filter CNNs \newline\cite{DBLP:journals/corr/abs-1711-07289} &
            Exact & Equivar. & $\SE(2)$ & \hl{Discretized} \newline (any angle)
            \\ \hline
        Learning invariance with weak supervision \newline\cite{Coors2018VISAPP} &
            \hl{Learned} \newline\tiny (Soft \newline constraints) & Invariance & $\SE(2)$ & Continuous  \\ \hline
        Roto-Translation Covariant CNNs \cite{DBLP:journals/corr/abs-1804-03393} &
            Exact & Invar. & $\SE(2)$ & \hl{Discretized} \newline (any angle)
            \\ \hline
        RotDCF: Decomposition of Convolutional Filters
        \newline\cite{DBLP:journals/corr/abs-1805-06846} &
            Exact & Equivariance & $\SE(2)$ & \hl{Discretized} \newline (any angle)\\ \hline
		RiCNN \newline\cite{RiCNN} &
		    Exact & Invar. & SO(2) & \hl{Discretized} (any angle) \\ \hline
        Siamese~Equivariant~Embedding~\mbox{\cite{DBLP:journals/corr/abs-1809-07217}} &
            \hl{Learned} \newline\tiny (Soft \newline constraints) & Equivar. & $\SO(2)$ & Continuous  \\ \hline
        CNN model of primary visual cortex
        \newline\cite{ecker2018a} &
            Exact &
            Equivariance & $\SE(2)$ & \hl{Discretized} \newline (any angle)\\ \hline
        General Steerable CNNs \cite{weiler2019general} &
            Exact &
            Equivar. & $\SE(2)$ & Continuous \\ \hline
    \end{tabularx}
    \caption{\label{tbl:equiv:2d}
    Methods with equivariance under 2D rotations.
    The terminology is summarized in Section~\ref{sec:equivariant-methods}.
    The input to each method is an image. In Polar Transformer Networks, the image is transformed to a circular signal in an intermediate layer.
    Methods with identical cell entries differ in terms of details.
    Potential weaknesses are \hl{highlighted}.
    General Steerable CNNs and their implementation in the \texttt{e2cnn} \protect\cite{weiler2019general} library are the ``best'' in that they provide various hyperparameter choices, with the other exact methods being special cases thereof (see Section~\ref{sec:equivariant-conv}).}
\end{threeparttable}
\end{table}
    \begin{table}[t!]
\setlength\arrayrulewidth{.1pt}
	\footnotesize
	 \begin{tabularx}{\columnwidth}{@{}|>{\raggedright\arraybackslash}X|L{0.79cm}|L{1.17cm}|L{0.69cm}|L{0.80cm}|L{1.58cm}|@{}}
		\hline
		Method & \hyperref[par:input]{Input} & \hyperref[par:approach]{Approach} & \hyperref[par:property]{Property} & \hyperref[par:group]{Group} & \hyperref[par:cardinality]{Cardinality}  \\
		\hline \hline
        Many & * &
            \hl{Learned} \newline\tiny (Data \newline augmentation) & 
            * & * & *  \\ \hline 
        Spatial Transformer Networks \cite{Jaderberg:2015:STN:2969442.2969465} & Voxel \newline grid &
            \hl{Learned} \newline\tiny (Learned~pose \mbox{normalization)} & Invariance & $\SE(3)$ & Continuous  \\ \hline
		Equivariant~Representations \cite{DBLP:journals/corr/abs-1711-06721} &
		    Spherical \newline signal & Exact &
		    Equivariance & $\SO(3)$ & Continuous  \\ \hline
		Spherical CNNs \newline\cite{DBLP:journals/corr/abs-1801-10130}&
		    Spherical~s. & Exact & Equivar. & $\SO(3)$ & Continuous  \\ \hline
		Tensor Field Networks \cite{DBLP:journals/corr/abs-1802-08219} &
		    Point cloud & Exact &
		    Equivariance & $\SE(3)$ & Continuous  \\ \hline
		N-body Networks \newline\cite{DBLP:journals/corr/abs-1803-01588} &
		    Point cloud & Exact &
		    Equivar. & $\SO(3)$ &
		    Continuous  \\ \hline
		CubeNet \cite{DBLP:journals/corr/abs-1804-04458} & 
		    Voxel grid & Exact & Equivar. & $\SE(3)$ & \hl{Discretized} \newline ($90^{\circ}$ angles)  \\ \hline
		3D G-CNNs \cite{3D-G-CNNs} &
	        Voxel grid & Exact & Equivariance & SE(3) & \hl{Discretized} ($90^{\circ}$/$180^{\circ}$ angles) \\ \hline
		3D Steerable CNNs \newline\cite{DBLP:journals/corr/abs-1807-02547} &
		    Voxel grid & Exact &
		    Equivar. & $\SE(3)$ & Continuous  \\ \hline
		PPF-FoldNet \newline\cite{DBLP:journals/corr/abs-1808-10322} &
		    Point cloud & Exact \tiny{(\hl{handcrafted features})} & Invar. & $\SE(3)$ & Continuous  \\ \hline
		Gauge Equivariant Mesh CNNs \cite{DBLP:journals/corr/abs-2003-05425} &
		    Polygon mesh & Exact & Equivariance & $\SE(3)$ & Continuous  \\ \hline
		$\SE(3)$-Equivariant DL for dMRI \cite{mueller2021} &
		    dMRI (6D) & Exact & Equivariance & $\SE(3)$ & Continuous  \\ \hline
	\end{tabularx}
	\caption{\label{tbl:equiv:3d}
        Methods with equivariance under 3D rotations.
        The terminology is summarized in Section~\ref{sec:equivariant-methods}.
        Potential weaknesses are \hl{highlighted}.
        Tensor Field Networks are the ``best'' for point clouds in that they provide continuous exact $\SE(3)$-equivariance.
        Similarly, 3D Steerable CNNs are the ``best'' neural networks for voxel grids.
        On the other hand, CubeNet and 3D G-CNNs offer only discrete rotations but are compatible with nonlinearities such as ReLU.
        The exact methods for 3D data are available via the \texttt{e3nn} \protect\cite{e3nn} library.
		}
\end{table}
    \begin{table}[t!]
\begin{threeparttable}[b]
\setlength\arrayrulewidth{.1pt}
	{\footnotesize
		\begin{tabularx}{\columnwidth}{@{}|>{\raggedright\arraybackslash}X|L{8mm}|L{9mm}|L{0.80cm}|L{10mm}|L{13.4mm}|@{}}
			\hline
			Method &
			\hyperlink{sec:rotout:input}{Input} &
			\hyperlink{sec:rotout:specialized}{Specialization} &
			\hyperlink{sec:rotout:group}{Group} &
			\hyperlink{sec:rotout:representation}{Embedding} & 
			\hyperlink{sec:rotout:loss}{Loss \newline function}
			\\ \hline \hline

			PoseNet \newline \cite{PoseNet} &
            Image &
            One object &
            $\SE(3)$ &
            Quaternion &
            $L^2$ dist.~in \hl{embedding space}
			\\ \hline

			Relative camera pose estimation using CNNs 
			\cite{RelativeCameraPoseEstimationUSingCNNs} &
            Video \tiny \newline (two non-con-secutive frames) &
            Generalizing to new objects &
            $\SE(3)$ &
            Quaternion &
            $L^2$ distance in \hl{embedding space}
			\\ \hline

			3D pose regression using CNNs \cite{3DPoseRegressionUsingCNNs} &
            Image &
            Multiple objects &
            $\SO(3)$ &
            Axis-angle~or quaternion &
            Geodesic distance
			\\ \hline

			Real-time seamless single shot 6D object pose prediction \newline \cite{SingleShot6DObjectPosePrediction} &
            Image &
            Multiple objects &
            $\SE(3)$ &
            Bounding box &
            Squared $L^2$ dist.~in \hl{embedding space}
			\\ \hline

			Registration of a slice to a predefined volume \newline \cite{DeepPoseEstimationWithGeodesicLoss} &
            Slice of volume &
            One object &
            $\SE(3)$ &
            Axis-angle representation &
            Geodesic distance\tnote{1}
			\\ \hline

			Registration of a volume to another, predefined volume \cite{DeepPoseEstimationWithGeodesicLoss} &
            Volume &
            One object &
            $\SE(3)$ &
            Axis-angle representation &
            Geodesic distance\tnote{1}
			\\ \hline

			SSD-AF \newline \cite{Egocentric6DOFTracking} &
            Crop-ped stereo image &
            Multiple objects &
            $\SE(3)$ &
            Vari-ous\tnote{2} &
            Smoothed $L^1$ dist.~in \hl{embedding space}
			\\ \hline

			Learning local~RGB-to-CAD~correspondences
			\cite{LearningRGBToCADCorrespondences} &
            Image and 3D model &
            Multiple objects &
            $\SE(3)$ &
            Rotation matrix &
            Squared $L^2$ dist.~in \hl{embedding space}
			\\ \hline

		\end{tabularx}
        \begin{tablenotes}
            {
            \item[1] Initially squared $L^2$ distance in embedding space (fast to compute); then geodesic distance for rotation and squared $L^2$ distance for translation.
            \item[2] Each method from the SSD-AF family uses a different embedding: discrete bins, four keypoint locations in 3D space, quaternion, Euler angles.
            }
        \end{tablenotes}
    	\caption{
    	    \label{tbl:output:gt}
    		Examples of deep learning methods that can output a 3D rotation, where a ground truth rotation is used for training.
            The terminology is summarized in Section~\ref{sec:rotout}.
            Losses that lack rotational invariance are \hl{highlighted}.
    	}
    }
\end{threeparttable}
\end{table}
    \begin{table}[t!]
\begin{threeparttable}[b]
\setlength\arrayrulewidth{.1pt}
	{\footnotesize
		\begin{tabularx}{\columnwidth}{@{}|>{\raggedright\arraybackslash}X|L{8mm}|L{9mm}|p{0.80cm}|L{10mm}|L{10mm}|@{}}
			\hline
			Method &
			\hyperlink{sec:rotout:input}{Input} &
			\hyperlink{sec:rotout:specialized}{Specialization} & 
			\hyperlink{sec:rotout:group}{Group} &
			\hyperlink{sec:rotout:representation}{Embedding} & 
			\hyperlink{sec:rotout:loss}{Loss \newline function}
			\\ \hline \hline

			Capsule Networks \newline \cite{DBLP:conf/nips/SabourFH17} &
            Image &
            Generalizing to new objects &
            $\SE(3)$ &
            Transformation~ma-trices\tnote{1}
            &
            \hl{Object classification}
			\\ \hline

			Spatial Transformer Networks \newline \cite{Jaderberg:2015:STN:2969442.2969465} &
            Volume &
            Generalizing to new objects &
            $\SE(3)$ &
            Transformation~matrix &
            \hl{Object classification}
			\\ \hline

			Unsupervised learning of depth and ego-motion
            \newline \cite{DepthAndEgoMotionFromVideo} &
            Video \tiny \newline (three consecutive frames) &
            Generalizing to new objects &
            $\SE(3)$ &
            Euler angles &
            View \newline warping
			\\ \hline

			Learning implicit representations of 3D object orientations from RGB \cite{ImplicitRepresentationsOfOrientations} &
            Image &
            One object
            &
            $\SO(3)$ &
            Learned representation &
            Auto-encoder
			\\ \hline

			GeoNet \newline \cite{GeoNet} &
            Video \tiny \newline (several consecutive frames) &
            Generalizing to new objects &
            $\SE(3)$ &
            Euler angles &
            View \newline warping
			\\ \hline

		\end{tabularx}
        \begin{tablenotes}
            \item[1] Poses of lowest-level object parts: learned representation; part-to-object pose transformations: transformation matrices (as trainable parameters).
        \end{tablenotes}
    	\caption{
    	    \label{tbl:output:nogt}
    		Examples of deep learning methods that can output a 3D rotation, where a ground truth rotation is \emph{not} necessary for training.
            The terminology is summarized in Section~\ref{sec:rotout}.
            Losses are \hl{highlighted} for which a good loss value does not guarantee a good prediction of rotations.
    	}
    }
\end{threeparttable}
\end{table}
		
		\item Volumetric data.
		The task is to estimate the orientation of an object relative to volume coordinates.
		\item Slice of volumetric data.
		The task is to estimate the orientation of a 2D slice relative to an entire (predefined) 3D object.
		\item Video:
		Two
		or three
		or more
		images (video frames). The task is to estimate the relative rotation and translation of the camera between (not necessarily consecutive) frames.
	\end{itemize}

	\item \hypertarget{sec:rotout:number}{Number of objects/rotations: Describes how many rotations the network outputs.}
	\begin{itemize}
		\item One rigid object:
		One rotation is estimated that is associated with a rigid object.
		The visible ``object'' is the entire scene in cases where camera motion relative to a static scene is estimated.
		Other small moving objects can be additionally accounted for (for example to refine the optical-flow estimation), but their \emph{rotation} is \emph{not} estimated.
		
		\item Hierarchy of object parts:
		Relative rotations between the objects and their parts are estimated, with several hierarchy levels, i.e.~an ``object'' consisting of parts can itself be one of several parts of a ``higher-level'' object. Among the methods listed in Tables~\ref{tbl:output:gt}--\ref{tbl:output:nogt}, only capsule networks belong to this category.
	\end{itemize}
	
	\item \hypertarget{sec:rotout:specialized}{Specialization:}
	\begin{itemize}
		\item Specialized on one object:
		The network can only process one type of object on which it was trained.
		\item Specialized on multiple objects:
		The network can process an object from an arbitrarily large but fixed set of object types on which it was trained.
		\item Generalizing to new objects:
		The network can generalize to unseen types of objects.
	\end{itemize}
	
	\item \hypertarget{sec:rotout:group}{Group: $\SO(3)$ or $\SE(3)$}
	
	\item \hypertarget{sec:rotout:representation}{Representation (embedding) of the rotation(s):}
	\begin{itemize}
		\item Rotation matrix
		\item Quaternion
		\item Euler angles
		\item Axis-angle representation
		\item Discrete bins:
		Rotations are grouped into a finite set of bins.
		\item Transformation matrix
		\item 3D coordinates of four keypoints on the object (predefined object-specifically, e.g.~four of its corners)
		\item Eight corners and centroid of 3D bounding box projected into 2D image space
		\item Learned representation:
		The latent space (e.g.~of an autoencoder) is used to represent the rotation.
		There are several interesting aspects at play:
		\begin{itemize}
			\item Learned representations allow for ambiguity:
			If an object looks very similar from two angles and the loss allows for it, the network can learn to use the same encoding to represent both rotations.
			On the other hand, unambiguous representations (like the ones listed above) would require generative/probabilistic models to deal with ambiguity.
			\item Certain representations are encouraged due to the overall network architecture.
			For example, in capsule networks, learned representations of rotation are processed in a very specific way (multiplied by learned transformation matrices).
			\item Features other than rotation might be entangled into the learned representation. This is not even always discouraged.
			For example, in capsule networks, the learned representation may also contain other object features such as color.
		\end{itemize}
	\end{itemize}

	\item \hypertarget{sec:rotout:loss}{Loss function. We distinguish the following categories:}
	\begin{itemize}
		\item Rotations are estimated at the output layer. The loss measures the similarity to ground truth rotations of training samples. These methods are listed in \autoref{tbl:output:gt}.
		
		\begin{itemize}
			\item Geodesic distance between prediction and ground truth:
			This loss is rotationally invariant, i.e.~the network miscalculating a rotation by $10^\circ$ always results in the same loss value, regardless of the ground truth rotation and of the direction into which the prediction is biased.
			\item $L^p$ distance in embedding space:
			This loss value is fast to compute but not rotationally invariant, 
			i.e.~an error of $10^\circ$ yields different loss values depending on the ground truth and on the prediction.
			Due to this ``unfairness''/``arbitrarity'', such losses are \hl{highlighted} in the table.
		\end{itemize}
		
		\item Rotations are estimated in an intermediate layer and used in subsequent layers for a ``higher-level'' goal of a larger system. Ground truth rotations are not required. These methods are listed in \autoref{tbl:output:nogt}.
		\begin{itemize}
			\item Object classification:
			Rotation prediction is trained as part of a larger system for object classification. The estimated rotation is used to rotate the input or feature map (in spatial transformer networks) or predicted poses (in capsule networks) as an intermediate processing step. It is assumed that learning to rotate to a canonical pose (in spatial transformer networks) or to let object parts vote about the overall object pose (in capsule networks) is beneficial for object classification. The estimation of rotation is incidental and encouraged by the overall setup.
			However, its approximate correctness
			is not necessary for perfect object classification. Therefore, the ``predicted rotation'' can be very wrong, and due to this danger this loss is \hl{highlighted} in the table.
			\item View warping:
			At least two video frames are used to estimate the scene geometry (depth maps) and camera motion between the views (rotation, translation). These estimates are used to warp one view (image, and possibly depth map) to resemble another view. The loss measures this resemblance. This is a form of self-supervised learning: ground truth geometry and motion are not given, but are estimated such that they cause warping that is consistent with the input images. The rotation estimation can be expected to be good, because it is necessary for good view synthesis.
			\item Autoencoder reconstruction loss:
			The network is trained to reconstruct its input (a view of the object) after passing it through a lower-dimensional latent space.
			The output target has a neutral image background and lacks other objects that were visible in the input image. This allows the network to learn to discard the information about the background and other objects before the bottleneck layer.
			If the network is specialized on one object, then maintaining in the latent space only the information about the object pose is sufficient for such reconstruction.
			If additionally the latent space is sufficiently low-dimensional, then the learning is encouraged to be economic about the amount of information encoded in the latent space, i.e.~to encode nothing but the pose.
		\end{itemize}
	\end{itemize}
\end{itemize}

Estimation of 2D rotations is simpler in terms of representation. Predicting the sine and cosine of the rotation angle (and normalizing the predicted vector to length $1$, because otherwise the predicted sine and cosine might slightly contradict each other, or be beyond $[-1,1]$) is better than predicting the angle, because the latter requires learning a function that has a jump (from $360^\circ$ to $0^\circ$), which is not easy for (non-generative) neural networks.

\subsection{Rotations as Input or as Deep Features}
\label{sec:input}
Other uses of rotations in deep learning are to take rotations as input, or to restrict deep features to belong to $\SO(n)$ (without requiring them to directly approximate rotations present in the data). For example, \cite{DBLP:journals/corr/HuangWPG16} use rotation matrices as inputs and as deep features. They restrict deep features to $\SO(3)$ by using layers that map from $\SO(3)$~to~$\SO(3)$.

\section{Conclusions}
\label{sec:conclusions}
Among the methods for equivariance/invariance, the most successful ones are based on the exact and most general approach (Section~\ref{sec:equivariant-conv}). They are very effective in 3D input domains as well.
With emerging theory~\cite{Kondor2018OnTG,DBLP:journals/corr/abs-1811-02017} for exact equivariance and with emerging approaches, it appears to be the perfect time to use the methods in various application domains and to tune them.
Existing pipelines that do not have (exact) equivariance yet and for example rely on data augmentation are likely to benefit from incorporating exact-equivariance approaches.

\section*{Acknowledgments}
We thank Erik Bekkers, Maurice Weiler, Gabriele Cesa, Antonij Golkov, Taco Cohen, Christine Allen-Blanchette, Qadeer Khan, Philip M\"uller, Philip H\"ausser, and Remco Duits for valuable discussions.
This manuscript was supported by the ERC Consolidator Grant ``3DReloaded'', the Munich Center for Machine Learning (Grant No. 01IS18036B), and the BMBF project MLwin.

\bibliographystyle{named}
\bibliography{refs}

\end{document}